\documentclass[sigconf]{acmart}

\AtBeginDocument{%
  \providecommand\BibTeX{{%
    \normalfont B\kern-0.5em{\scshape i\kern-0.25em b}\kern-0.8em\TeX}}}

\setcopyright{acmlicensed}
\copyrightyear{2018}
\acmYear{2018}
\acmDOI{XXXXXXX.XXXXXXX}

\acmConference[KDD'24]{}{August 25--29,
  2024}{Barcelona, Spain}
\acmISBN{978-1-4503-XXXX-X/18/06}

\usepackage{makecell}
\usepackage{multirow}
\usepackage{algorithm}
\usepackage{algpseudocode}
\usepackage{tabularx}
\usepackage{graphicx}
\usepackage{enumitem}
\usepackage{multicol}

\newcounter{savealgorithm}

\makeatother
\algnewcommand\algorithmicforeach{\textbf{for each}}
\algdef{S}[FOR]{ForEach}[1]{\algorithmicforeach\ #1\ \algorithmicdo}

\algdef{SE}[REPEATN]{RepeatN}{End}[1]{\algorithmicrepeat\ #1 \textbf{times}}{\algorithmicend}



\newenvironment{compact_enum}{
\begin{itemize}
  \setlength{\itemsep}{-0pt}
  \setlength{\parskip}{-0pt}
  \setlength{\parsep}{-0pt}
  \setlength{\itemindent}{-0pt}
  \labelsep=12pt
}{\end{itemize}}

\title{LinkSAGE: Optimizing Job Matching Using \\ Graph Neural Networks}

\author{Ping Liu, Haichao Wei, Xiaochen Hou, Jianqiang Shen, Shihai He, Kay Qianqi Shen, Zhujun Chen, Fedor Borisyuk, Daniel Hewlett, Liang Wu, Srikant Veeraraghavan, Alex Tsun, Chengming Jiang, Wenjing Zhang}
\affiliation{%
  \institution{LinkedIn Corporation}
  \city{Mountain View}
  \state{CA}
  \country{USA}
  \postcode{94043}
}
\email{{piliu, hawei, xiahou, jershen, she1, qishen, zhuchen, fborisyuk, dhewlett, liawu, atsun, cjiang, wzhang}@linkedin.com}

\begin{abstract}

We present LinkSAGE, an innovative framework that integrates Graph Neural Networks (GNNs) into large-scale personalized job matching systems, designed to address the complex dynamics of LinkedIn's extensive professional network. Our approach capitalizes on a novel job marketplace graph, the largest and most intricate of its kind in industry, with billions of nodes and edges. This graph is not merely extensive but also richly detailed, encompassing member and job nodes along with key attributes, thus creating an expansive and interwoven network.
A key innovation in LinkSAGE is its training and serving methodology, which effectively combines inductive graph learning on a heterogeneous, evolving graph with an encoder-decoder GNN model. This methodology decouples the training of the GNN model from that of existing Deep Neural Nets (DNN) models, eliminating the need for frequent GNN retraining while maintaining up-to-date graph signals in near real-time, allowing for the effective integration of GNN insights through transfer learning. The subsequent nearline inference system serves the GNN encoder within a real-world setting, significantly reducing online latency and obviating the need for costly real-time GNN infrastructure.
Validated across multiple online A/B tests in diverse product scenarios, LinkSAGE demonstrates marked improvements in member engagement, relevance matching, and member retention, confirming its generalizability and practical impact.

\end{abstract}

\begin{document}
\pagestyle{empty} 
\maketitle
\thispagestyle{empty} 

\section{Introduction}

LinkedIn, the world’s largest professional networking platform with over 1B members, serves as a dynamic environment for career development, job recruitment, and professional growth. The platform connects individuals based on commonalities like educational backgrounds and professional experiences, fostering a network where opportunities and competencies are enhanced through these connections. On average, our job product service, including recommendation and search, facilitates over 85M job applications on our platform every week.
However, job matching on LinkedIn presents unique challenges. The platform, with its extensive data on job opportunities and members, faces various complexities:

\begin{compact_enum}
\item {\textbf{Dynamic Nature of Jobs}}: Jobs on LinkedIn are dynamic and short-lived. This fast-paced environment demands timely relevance in matching jobs to potential candidates, a more intricate task than in other recommender systems.
\item {\textbf{Sparse Engagement}}: The interaction between members and job listings is often sparse and temporal, complicating engagement analysis.
\item {\textbf{Multidimensional Matching}}: Job matching involves diverse factors including skills, education, industry experience, and seniority. Real-world scenarios often see partial matches leading to successful job placements.
\item {\textbf{Cold-Start Problem}}: Addressing the cold-start issue is crucial, especially for users with limited experience or those seeking career changes, and for newly posted jobs with minimal activity.
\item {\textbf{Large-Scale Evolving Data}}: The platform's vast user base, exceeding 1 billion members, and millions of jobs and companies, contribute to its large-scale   heterogeneous data.
\end{compact_enum}

Graph Neural Networks (GNN) have emerged as a promising solution to these challenges. GNNs are adept at handling dynamic relationship in real-time, managing complex relational datasets, and representing diverse data types like skills, geographies, and industries as node types. Their message-passing capabilities effectively mitigate the cold-start problem and can provide more relevant job-matching results.

\begin{figure}
    \centering
    \includegraphics[width=2.7in]{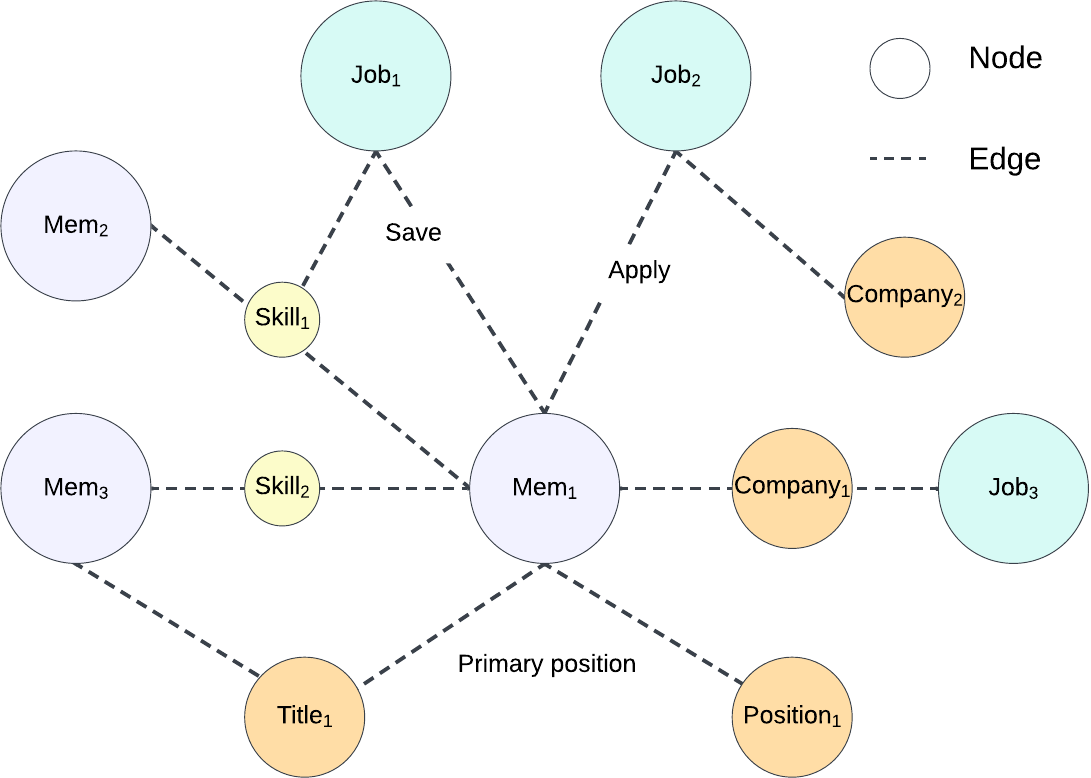}
    \caption{An example of LinkedIn's job marketplace graph. Members (also referred to as job seekers) are connected to their attributes, including Skill, Title, Company, and Position. The same connection is applied for each job. A member is connected to a job as well if there are engagements happening (e.g., save, apply, or click).}
    \label{fig:marketplace}
\end{figure}

While GNNs have been implemented in various large-scale recommender systems, integrating them into existing product systems, often powered by deep neural networks (DNNs), remains a challenge. 
This paper proposes an industrial framework for applying GNN within large-scale recommender systems that are currently driven by DNNs. This framework is not only economically feasible but also ensures a non-disruptive transition to a GNN-based model.
Following this framework, we have developed the largest job marketplace graph in the industry, with billions of nodes and edges, including members, jobs, and essential attributes. This graph is a pioneering effort in the field, as shown in Figure~\ref{fig:marketplace}. Our unique contributions include:

\textit{\textbf{GNN Training Methodology}}: Our framework employs a novel approach by integrating inductive graph learning on a heterogeneous, evolving graph with an encoder-decoder GNN model. This approach avoids the needs of frequently retraining of GNN model (which is practically impossible due to the evolving nature of the large scale, dynamic graph) while keeping the graph always up to date in near real-time for training and serving the downstream DNN models. It distinctly separates the training of a dedicated encoder-decoder GNN model from the training of the pre-existing DNN models in a supervised setting. This sequential training process first involves training the GNN encoder, which is then incorporated into existing DNN models. This method facilitates the seamless incorporation of GNN insights into established DNN models through transfer learning, resulting in significant enhancements with minimal impact on online latency.

\textit{\textbf{GNN Serving via Nearline Inference}}: The framework introduces an innovative serving mechanism for the GNN encoder, incorporating it into existing DNN models within the recommender system through a nearline inference pipeline. This strategy involves precomputing the computationally intensive GNN encoder and storing its output in an in-memory feature store, which is then utilized by the DNN models. This approach effectively eliminates the need for expensive real-time GNN infrastructure and achieves significant improvements in relevance performance, while ensuring latency remains in the low tens of milliseconds range. The nearline inference system is particularly vital due to the vast number of jobs posted on LinkedIn daily. Without this system, the full potential of the GNN would not be realized until the next day's inference task is complete, a challenge we address in Section \ref{sec.training}.

\textit{\textbf{Relevance Matching on Equality and Segments}}: The designed heterogeneous graph enables the information propagation through edges, allowing member nodes with less training data to receive significant information from its neighboring nodes that have more robust data during the training of our GNN encoder-decoder model. Our application of GNN encoder to existing neural networks has surprisingly improved relevance matching across all of member base from power members to infrequent visitors historically lacking in predictive data. This means a node with less training data can still receive significant information from its neighboring nodes that have more robust data.


By deploying LinkSAGE to production, our work not only showcases the generalization and effectiveness of LinkSAGE in applying GNN to improve various business metrics across different use cases, such as member engagement and relevance matching, but also demonstrates its potential in promoting equality and inclusivity in job recommendations. The same method can be applied to many recommender systems.


The rest of paper is organized as follows. We discuss the related work in Section \ref{sec.related_work}. We describe our graph definition in Section \ref{sec.graph_definition}. We explain our training settings in Section \ref{sec.training}, illustrate how to integrate GNN into online models in Section \ref{sec.model_serving}, then discuss and analyze our online experimental results in Section \ref{sec.experiment}. We conclude our work and discuss future work in Section \ref{sec.conclusion}.

\section{Related Work}
\label{sec.related_work}

Graph neural networks (GNNs) have demonstrated impressive capabilities in handling structured and relational datasets across various domains, including social networks, biology, chemistry, recommender systems, and computer vision \cite{wu2020graph,fout2017protein,do2019graph,ying2018graph,satorras2018few}. Graph Convolutional Networks (GCNs) \cite{kipf2016semi, defferrard2016convolutional} stand out for learning shared filter parameters from graph data. Concurrently, the GraphSAGE framework \citet{hamilton2017inductive} offers flexible inductive aggregation options to manage dynamic neighbors in graphs. Furthermore, Graph Attention Networks (GATs) \cite{velivckovic2018graph} utilize masked self-attentional weights across various neighbors, indicating a trend towards more complex graph structures \cite{feng2019hypergraph,zhang2019heterogeneous,yang2023simple}.

In the industry, GNN applications are predominantly seen in recommender systems. Pinterest \cite{ying2018graph} pioneered the use of large-scale web-based graph neural networks for recommendations. Twitter \cite{el2022twhin,tgn_icml_grl2020} developed and open-sourced the TwHIN framework, which employs pre-trained entity representations to enhance downstream recommendations, achieving notable offline and online results. Uber \cite{bose2019meta} also utilized pre-trained GNN representations in candidate selection and personal ranking models. Google Maps has recently implemented large-scale spatio-temporal interactions in its ETA predictions using these techniques \cite{derrow2021eta}.

The job recommender system typically mirrors the methodologies of general recommender systems \cite{de2021job}. Content-based \cite{kessler2012hybrid,mpela2020mobile} and collaborative filtering \cite{lee2017exploiting,reusens2017note} approaches are popular in academic research. Knowledge-based systems \cite{gutierrez2019explaining}, which focus on matching and similarity in an ontology space \cite{de2021job,rivas2019hybrid,martinez2020novel}, are unique to job recommendation. Recently, hybrid systems and deep neural networks have gained traction \cite{nigam2019job}. \citet{qin2018enhancing} leveraged rich text data from job requirements and seekers' experiences, using Recurrent Neural Networks for person-job fit prediction. \cite{le2019towards} introduced an interpretable job-seeker fit system, providing reasoning behind recommendations.

The pursuit of equity in recommender systems has gained significant attention \cite{10.1145/3547333}, with a growing body of research dedicated to ensuring equitable outcomes for all users \cite{lee2019procedural,ge2021towards}. The issue of data sample bias in machine learning-based systems, stemming from various sources such as data or algorithmic models, presents a notable challenge \cite{chen2023bias}. Ensuring balanced representation and fair treatment across diverse user segments is a critical, ongoing area of focus \cite{liu2019personalized}. In Section \ref{sec.experiment}, we delve into the results of our A/B testing, specifically examining the system's performance in terms of equity and its impact on different user segments.

\section{Graph Construction}
\label{sec.graph_definition}

In this section, we discuss how we construct the job marketplace graph including node types and edge types. Following a neighborhood aggregation scheme, the job marketplace graph is designed to recursively compute the representation vector of a node by aggregating and transforming representation vectors of its neighboring nodes. 
The performance of our GNN model is tied to the efficient flow of information between entities and hence highly depends on the underlying graph's structure.
A high-quality, heterogeneous graph that encodes rich information can significantly enhance the performance of our models across the ranking system stack. We leverage our prior knowledge on hiring efficiency to design heuristic rules that guide the  graph construction.

In terms of node types, we introduce 6 categories of entities to the graph: member, job, skill, title, company, and position. Nodes for skills, titles, and companies are extracted attributes from members/jobs, while the position node represents a $\langle$company, title$\rangle$ tuple which provides a clearer definition of a job opening. Based on interviews with job seekers, recruiters, and thorough data analysis on hiring outcomes, we consider these factors pivotal in determining the quality of job matching. The detailed  statistics on those nodes are shown in Table~\ref{tab.stats.node}.

Introduced edges can be categorized into 2 groups. The first group links  members and jobs to their static attributes, with each edge type corresponding to a different attribute node. For instance, if a seeker is a ``software engineer'', we will link this seeker to title node ``\textit{software engineer}''; if a job is from company ``dream inc'', we will link this job to company node ``\textit{dream inc}''.
The second group relies on implicit connections between members and jobs, inferred from their interactions. We believe that if a member and job pair has interactions, it implies the existence of commonalities linking them together.
In our graph, edges from member to job nodes symbolize user engagement (e.g., saving, applying, clicking a job), while those from job to member nodes reflect recruiter interactions (e.g., reach out to seekers). To further augment graph connectivity, we introduce reciprocal edges manually.

Our training data set contains 1B members and 50M jobs. The initial version of our graph was very sparse and there were barely any connections between member$\Leftrightarrow$job, member$\Leftrightarrow$member, or job$\Leftrightarrow$job pairs. We tackled this issue by densifying the graph through the inclusion of skill connections. In recent years, LinkedIn has embarked on a journey to help our members and customers hire for the future, making skill-first investments to foster a more equitable and efficient job market. 
However, it's important to note that, on average, a member possesses around 17-18 skills, while a job typically lists 30+ skills as requirement, many of which may not be crucial to job matching. For instance, John may have skills such as \textit{HTTP}, \textit{Java}, and \textit{Machine Learning} in his profile, but he is likely primarily interested in jobs related to \textit{Machine Learning}.

We identify top skills by building a scoring model to assess the relevance of a particular skill to a seeker or a job. We then add links to connect these top skills to the corresponding seekers or jobs.
Skills play a critical role by facilitating the exchange of information between members with similar skill sets and enabling information flow from jobs to members (or vice versa) when a member acquires skills relevant to a particular job.
We have found that GNN, when properly designed to incorporate skill-level information, can foster meaningful connections between skills, members, and jobs, leading to enhanced model performance. Table. \ref{tab.stats.edge} shows that on average a member is linked to 1.2 top skill, and job is linked to 0.67 top skill in our graph.
The resulting graph is vast, encompassing billions of nodes and edges, with detailed statistics available in Table~\ref{tab.stats.edge}.

\begin{table}
\begin{center}
\begin{tabular}{lc|lc}
\toprule
Node Type & \# of Nodes &  Node Type & \# of Nodes \\
\midrule 
member  & 1B    & job      & 50M  \\
title   & 25K   & position & 195M \\
company & 25M   & skill    & 41K   \\
\bottomrule
\end{tabular}
\end{center}
\caption{Statistics of Nodes in LinkedIn job marketplace graph}
\vspace{-12pt}
\label{tab.stats.node}
\end{table}

\begin{table}
\begin{center}
\small
\begin{tabular}{@{\hskip 8pt}l@{\hskip 8pt}c|@{\hskip 8pt}l@{\hskip 8pt}c}
\toprule
Edge Type & \# of Edge & Edge Type & \# of Edge \\
\midrule
member-title          & 1B    & job-title       & 46M  \\
member-company        & 966M  & job-company     & 42M \\
member-position       & 139M  & job-position    & 41M   \\
member-skill          & 1.2B  & job-skill       & 33M   \\
recruiter interaction & 26M   & seeker engagement & 2.7B \\
\bottomrule
\end{tabular}
\end{center}
\caption{Statistics of Edges in LinkedIn job marketplace graph}
\vspace{-12pt}
\label{tab.stats.edge}
\end{table}

We believe that the effectiveness of GNN can be enhanced by leveraging skill connections to improve graph construction capabilities and by inferring skill signals.
Our offline GNN model evaluations, which included skill nodes and focused on extracting top skills relevant to both jobs and members, demonstrated a notable improvement. Specifically, we observed a $+1.5\%$ increase in recall in our offline models compared to the baseline models that lacked skill nodes in the job marketplace graph.

\section{Model Training}

In this section, we discuss how we prepare the graph data, model architecture. We also state how we design our model training in our production.

\label{sec.training}
\subsection{Data Preparation}
To train the GNN model, we need to construct both graph data and label data. The graph data construction is highly important and extensively discussed in Section \ref{sec.graph_definition}. Once the graph is formed, the next step is to employ a graph engine to store the graph data and provide real-time graph sampling information during GNN model training. Several products cater to this purpose, such as DeepGNN \cite{deepGNNref}, PyG \cite{Fey/Lenssen/2019}, and DGL \cite{wang2019dgl}. In our case, we opt for DeepGNN due to its scalability and compatibility, and it also supports various graph sampling algorithms including multi-hop random sampling, weighted sampling, and Personalized PageRank (PPR) sampling \cite{chien2021adaptive}. The graph data is stored in binary format in DeepGNN, so we need to utilize the data converter component provided by DeepGNN to transform the raw graph data into binary format. 

The label data preparation is closely related to how the job recommendation task is modeled. It's natural to formulate the job recommendation application as a link prediction problem, wherein the objective is to predict the existence of an edge between a member node and a job node. Consequently, the label data is structured as tuples in the form of (memberId, jobId, label), where the label field indicates whether to recommend the job to the member.

\subsection{Model Architecture}
\begin{figure*}[ht]
    \centering
    \includegraphics[width=6.9in]{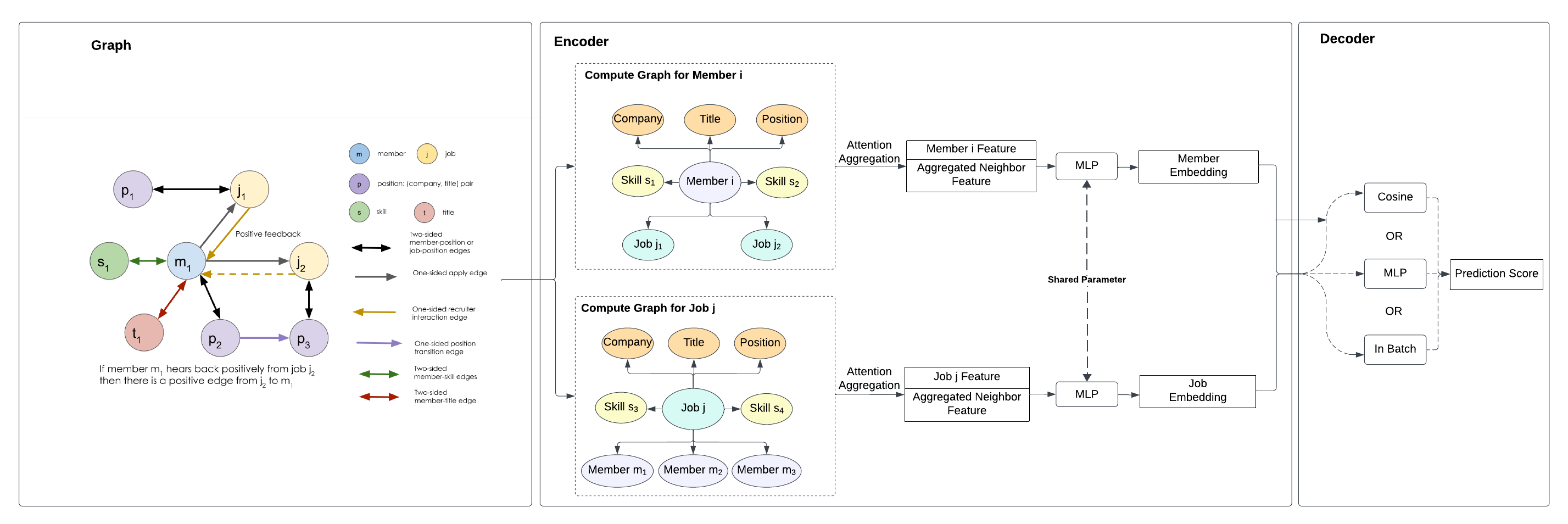} 
    \caption{Model Architecture}
    \label{fig:model_architecture}
\end{figure*}
As illustrated by Figure \ref{fig:model_architecture}, our Graph Neural Network  model adopts an encoder-decoder architecture.
The encoder, specifically, utilizes the GraphSAGE algorithm~\cite{hamilton2017inductive} for its ability to handle large-scale, evolving graphs inductively. This encoder works by aggregating information from the neighbors of a query node to generate its representation. 

Let  $\mathcal{M}$ be the matrix of member embeddings in a batch, $ \mathcal{J}$ be the matrix of job embeddings in a batch, and \( \text{score}(i, j) \) be the score for the \( i \)-th member and the \( j \)-th job.

Mean aggregation updates embeddings as:
\[ \mathcal{M}_i = \frac{1}{|\mathcal{N}(i)|} \sum_{n \in \mathcal{N}(i)} f(\text{features}(n)) \]
\[ \mathcal{J}_j= \frac{1}{|\mathcal{N}(j)|} \sum_{n \in \mathcal{N}(j)} f(\text{features}(n)) \]

Attention-based aggregation updates embeddings as 
\[ \mathcal{M}_i = \sum_{n \in \mathcal{N}(i)} \alpha(i,n) \times f(\text{features}(n)) \]
\[ \mathcal{J}_j = \sum_{n \in \mathcal{N}(j)} \alpha(i,n) \times f(\text{features}(n)), \]
\sloppy where \( \mathcal{N}(i) \) denotes the set of neighbors of node \( i \), \( f(\text{features}(n)) \) represents the feature transformation function, \( \alpha(i,n) \) are the attention scores between node \( i \) and its neighbor \( n \), computed as part of the attention mechanism.

In this manner, the encoder aggregates features from neighbors to produce the embedding matrices \( \mathcal{M}_i \) and \( \mathcal{J}_j \) for members and jobs, respectively.
Following the encoder, these embeddings are fed into the decoder to calculate prediction scores. Our model supports various decoders: the Multilayer Perceptron (MLP) decoder, the cosine decoder, and the in-batch negative sampling decoder.

Let's illustrate with the example of in-batch negative sampling. The prediction score between each member and job pair in the batch is calculated using the dot product of their embeddings, expressed as:
\[ \text{score}(i, j) = \mathcal{M}_i \cdot \mathcal{J}_j \]
where \( \mathcal{M}_i \) is the embedding for the \( i \)-th member and \( \mathcal{J}_j \) is the embedding for the \( j \)-th job.

The loss function for this model is the cross-entropy loss, which compares the predicted scores against the true labels. Given the member set $S_M$ and the job set $S_J$, the total loss over the training data is expressed as:

\begin{align*}
 \text{Loss} = -\sum_{i\in S_M}\sum_{j\in S_J} ( y_{ij} \log(\sigma(\text{score}(i, j))) \\
 + (1 - y_{ij}) \log(1 - \sigma(\text{score}(i, j)))),
\end{align*}
where \( \sigma \) is the sigmoid function and \( y_{ij} \) is the true label for the pair \( (i, j) \), which is 1 if the pair is a positive example and 0 otherwise.

\subsection{Training and Optimization}
\begin{figure}
    \centering
    \includegraphics[width=3in]{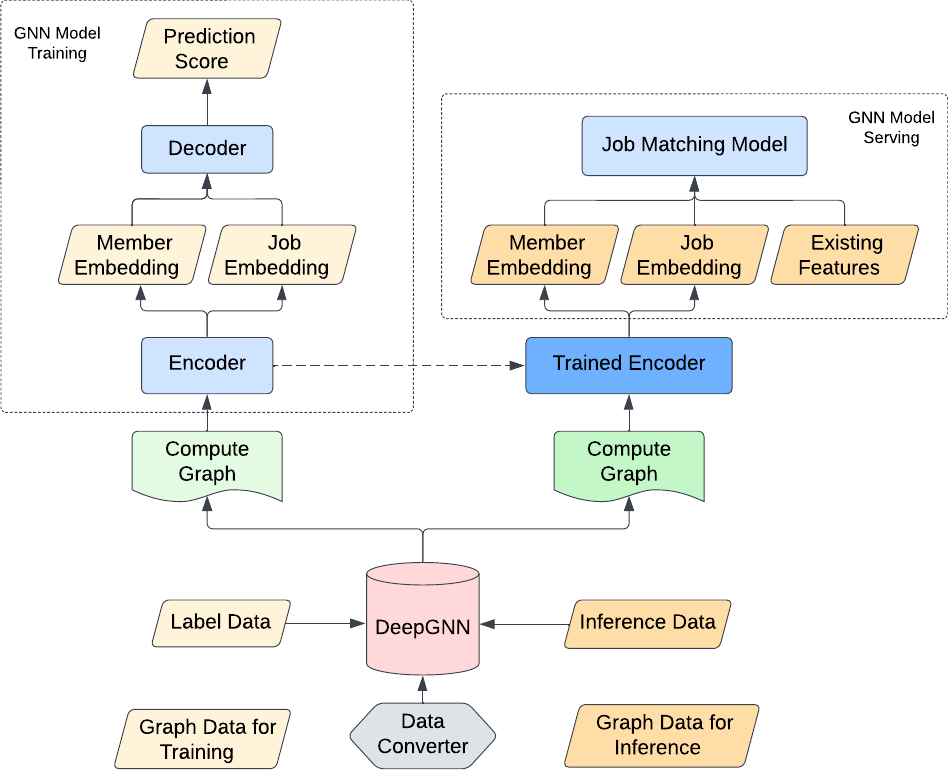}
    \caption{Model training and serving.}
    \label{fig:gnnModelTraining}
\end{figure}

The overall design of the model training pipeline is illustrated in the left section of Figure ~\ref{fig:gnnModelTraining}. The process begins with input label data, structured as tuples of $\langle$memberId, jobId, label$\rangle$. These tuples are then processed where memberId and jobId are utilized as query nodes in DeepGNN. DeepGNN, leveraging the existing graph data, constructs a compute graph centered around these query nodes. This compute graph comprises the query node (either a member or a job), neighbors sampled according to a predefined sampling algorithm, and their associated features.

During the iteration of training the model to get better performance, a key insight emerged: enhancing graph connectivity is crucial. This led to a focused effort on refining the design of the graph's edges. Our observations indicated that optimal performance was achieved when certain edge configurations were employed. Specifically, we found that creating bidirectional edges between members and their titles as well as their skills was most effective. Additionally, we maintained bidirectional edges for position transitions. Furthermore, we established bidirectional edges between both members and jobs to their respective positions. This structured approach in edge design as illustrated by the graph section of Figure \ref{fig:model_architecture} significantly contributed to the enhanced performance of our model by fostering a more interconnected and robust graph structure.


\section{Model Serving}
\label{sec.model_serving}

The GNN encoder, once trained, is integrated into the ranking models for serving online traffic, as illustrated in the right section of Figure ~\ref{fig:gnnModelTraining}. To improve personalization experience to our members, we built a nearline infrastructure to continuously update GNN embeddings. GNN serves as a platform for AI engineers, facilitating feature engineering and seamless integration of signals into our ecosystem. GNN has been served as a platform for our AI engineers to conduct feature engineering and integrate signals into our ecosystem.

\subsection{Integration with Ranking Models}
To meet stringent latency requirements in our online environment, the actual ranking models that is served online is a lighter DNN-based model by plugging in the well trained GNN encoder. The right part of Figure~\ref{fig:gnnModelTraining} illustrates the process. A job matching model concatenates the member and job embedding from the GNN encoder model with other relevant features and is trained with respective objective function. To avoid label leakage, we ensure that the GNN model's training and graph data predate those used for training the job matching model.

This approach eliminates the need for real-time computation of member and job embeddings. Instead, they can be precomputed offline and cached in a key-value store. In the offline inference flow, the most recent activities of both members and jobs are integrated into the graph data and used to update their embeddings. During the online inference, we retrieve them based on their respective IDs as inputs to the ranking model.
This significantly reduces online latency, from seconds to just tens of milliseconds. The implementation detail is described in Section \ref{sec.nearline_inference}.

The adaptability of this design extends to incorporating the trained GNN encoder into various downstream models through transfer learning. This approach not only showcases the flexibility of our model but also its capacity for generalization, leading to significant improvements in a range of online metrics across multiple applications. 


\subsection{Nearline Inference Framework}
\label{sec.nearline_inference}

\begin{figure*}[ht]
\centering
  \includegraphics[width=6.7in]{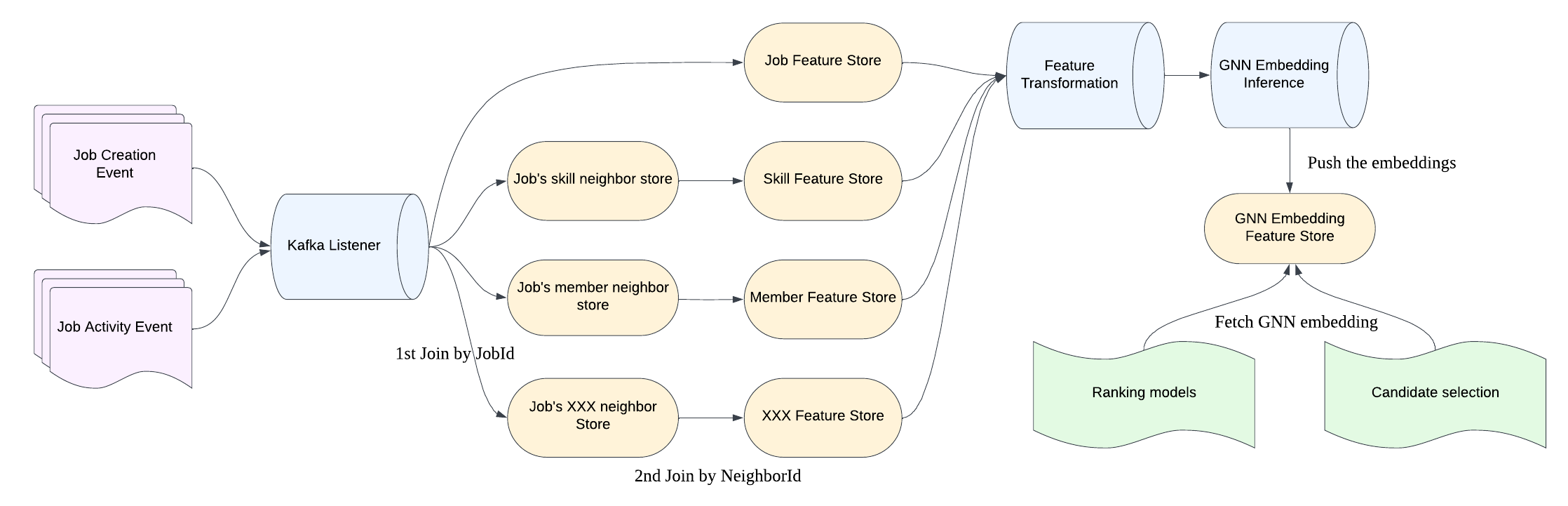}
  \caption{The working flow of nearline inference pipeline for job embeddings.}
  \label{fig:nearlineInference}
\end{figure*}

``\textit{Nearline}'' model inference, distinct from ``\textit{real-time}'', offers a more lenient time frame for event handling. Real-time model inference involves immediate computations, typically within milliseconds, while in the nearline system, there is typically an acceptable delay of a few seconds or even a few minutes. Our previous ``offline'' approach involved daily batch jobs to generate member and job embeddings and store them in the key-value store, leading to a 24-hour delay for newly created jobs to receive their GNN embeddings. This delay also affected the reflection of member engagement activities in the job marketplace graph. LinkedIn has a highly dynamic repository of job postings, each with an average lifespan of $\sim$3 weeks. 
Job seekers are active on our platform and over 50\% of them resume their job-seeking sessions within 10 minutes at least once per month. 
To enhance the experience for our members, we require more timely updates for embeddings, transitioning from offline to nearline processing.

Implementing a nearline GNN inference pipeline faces several challenges, particularly given that we still need the graph engine serving the query request of graph data in real-time. Key difficulties include:
(a) The job marketplace graph size ranges from 3-5TB, requiring equivalent physical memory for loading, which is costly.
(b) Real-time updates are operationally demanding due to their frequency.
(c) The graph engine's dependence on significant resources introduces I/O bottlenecks, resulting in increased latency -- a critical concern in industrial production.

To tackle these challenges, we introduced a nearline inference framework that utilizes sequential joining and an in-memory NoSQL database-based feature store. This approach creates a ``stateful'' job marketplace graph for production without the need for a real-time, fully operational graph engine. This streamlined solution is feasible and practical because, during inference, only the neighbors and their input features are required, rather than the full spectrum of capabilities like temporal processing and sampling typically associated with a comprehensive graph engine.

Taking job embedding as an example, Figure~\ref{fig:nearlineInference} illustrates the workflow for job embedding in our nearline inference system. The system is triggered in two scenarios: (1) when a recruiter creates a new job posting, and (2) when new neighbors (e.g., members who applied/saved/clicked on a job posting) are introduced to an existing job. These triggers are monitored via Kafka messagings. Neighbor data is continuously updated and stored in NoSQL-type feature stores keyed by job IDs. 
Our configuration includes multiple feature stores that monitor job neighbors per node type, accompanied by a secondary store for the input features of each neighbor.
By defining a sequential join in the NoSQL store, we fetch nodes, neighbors, and input features, perform necessary transformations, and feed them as features into the nearline GNN model inference flow. The resulting embeddings are pushed to an online feature store used as precomputed features for Job Matching Models (including job ranking and candidate selection models).


This nearline inference system, currently operational in our production environment, notably reduces latency to tens of milliseconds, with a peak QPS (Query per Second) exceeding 5K. It serves as our foundation for feature engineering effort and we observed notable business improvements with the implementation of this system.

\section{GNN as a Platform for Feature Engineering}

Our vision for GNN training and deployment aims to enhance our matching capabilities into a cohesive "job marketplace stack" at LinkedIn. 

We maintain a variety of ranking models to connect members with opportunities like jobs, recruiters, and courses. We've found that certain signals—particularly those reflecting a member's experience and expertise—are universally beneficial across these models. Traditionally, the incorporation of new signals into our system has been gradual, often beginning with a single model before extending to others. To improve AI productivity, we implement GNN as a new horizontal-first working model where multiple signals can be incorporated rapidly across the full suite of AI models for our Job Marketplace.

We have leveraged GNN infrastructure to seamlessly introduce new signals into the marketplace's ranking models, marking a pivotal shift in our development strategy. This shift prioritizes GNN for the initial integration of signals, with AI engineers across various teams contributing. By employing a development model akin to git branching, engineers have the flexibility to independently explore and assess the impact of new signals on their specific models by branching out GNN model. These efforts are then merged into a main GNN model branch, which, once put into production, necessitates the retraining of downstream models to harness these advancements fully.

The GNN framework proved highly effective in our signal integration workflow. By integrating signals into GNN as a primary step, engineers from varied teams successfully achieved significant improvements in a wide array of applications. This cross-functional collaboration facilitated through GNN not only streamlined the integration process but also amplified the overall impact on our product suite. For an in-depth analysis of these advancements, please refer to Section \ref{sec.experiment}.

\section{Experiments}
\label{sec.experiment}
We evaluated the generalization and effectiveness of our GNN system  on various LinkedIn Jobs Marketplace downstream models through online A/B tests. The experiments focus on machine learning models for candidate selection and personalized ranking. Significant results are reported in comparison to the baseline, determined by a hypothesis t-test with a p-value less than 0.05 in two-tailed testing. Major metric definitions can be found in Table~\ref{tab.metrics_def}.

\begin{table*}
\begin{center}
\begin{tabular}{@{\hskip 8pt}l@{\hskip 8pt}l}
\toprule
Metric Name &  Definition \\
\midrule
Job Sessions                  & Number of jobs touching sessions as defined as jobs ecosystem in LinkedIn.       \\
Qualified Applications  (QA)  & Applications to jobs that have owners where the applicant was targeted or matched to jobs' attributes.   \\
Qualified Applications Rate   & The ratio between number of QA and number of total applies  \\
Dismiss To Apply Ratio        & The ratio between number of total dimiss actions and number to total apply clicks  \\
\bottomrule
\end{tabular}
\end{center}
\caption{Online Metrics Definition }
\vspace{-8pt}
\label{tab.metrics_def}
\end{table*}

\subsection{Top Applicant Jobs (TAJ)}
The Top Applicant Job (TAJ) model is designed for LinkedIn Premium members. It aims to increase recruiter interactions (emails, messages, phone calls) following job applications. We approach this by treating recruiter interactions prediction as a link prediction problem, utilizing inferred GNN encoded member representation for personalization in the TAJ ranking model.

Table \ref{tab.taj.1} presents the outcomes of incorporating GNN encoded member representation into the TAJ model. Data over a 12-week period reveals a notable increase in engagement among Premium members. We observed a rise in recruiter interaction metrics such as Positive Hearing Back rate, aligning with our optimization goals. Premium hence gained more value through their membership and were more willing to renew their membership.


\begin{table}
\begin{center}
\begin{tabular}{@{\hskip 20pt}l@{\hskip 8pt}l}
\toprule
Metric Name &  Impact \\
\midrule
Company Follows                          & $+1.8\%$   \\
Positive Hearing Back rate               & $+1.0\%$  \\
Onsite Application Positive Rating Rate & $+1.3\%$  \\
Renewal Rate                             & $+0.3\%$  \\
\bottomrule
\end{tabular}
\end{center}
\caption{Relative metrics of the first online A/B test in TAJ. }
\vspace{-8pt}
\label{tab.taj.1}
\end{table}

\begin{table}
\begin{center}
\begin{tabular}{@{\hskip 20pt}l@{\hskip 8pt}l}
\toprule
Metric Name &  Impact \\
\midrule
Survival Rate                          & $+2.2\%$   \\
Apply Clicks Positive Interaction      & $+20.0\%$  \\
\bottomrule
\end{tabular}
\end{center}
\caption{Relative metrics of the second online A/B test in TAJ. }
\vspace{-8pt}
\label{tab.taj.2}
\end{table}


In our second A/B test conducted a few months later, we updated the GNN model by enhancing the graph with more recruiter-member edges. Again, this led to  a significant rise in positive interaction between Premium job seekers and recruiters. 
A $+2.2\%$ relative increase in the survival rate (transition from free to paid subscription) was likely related to such additional value offered by our platform.

\subsection{Job You May Be Interested In (JYMBII)}
Jobs You May Be Interested In (JYMBII) is LinkedIn’s primary personalized service to recommend jobs to member. We conducted a comprehensive A/B test to measure the job seeker engagement from top of funnel job seeking sessions to bottom of the funnel focusing on the volume of qualified applications.


Table \ref{tab.jymbii} displays the key improvements in the JYMBII model after incorporating GNN encoder. Notable enhancements include a $+2.2\%$ relative increase in Qualified Applications (QA) and a $+0.3\%$ relative rise in QA rate.
The growth in job session metrics reflects better relevance of top recommended jobs, while a decreased job Dismiss to Apply ratio indicates higher member satisfaction. This experiment demonstrates the substantial ecosystem-level impact of skill-based GNN model in the job marketplace.

\begin{table}
\begin{center}
\begin{tabular}{@{\hskip 20pt}l@{\hskip 8pt}l}
\toprule
Metric Name &  Impact \\
\midrule
Qualified Applications                 & $+2.2\%$   \\
Qualified Applications Rate            & $+0.3\%$   \\
Job Sessions                           & $+0.6\%$   \\
Dismiss To Apply Ratio                 & $-6.0\%$   \\
\bottomrule
\end{tabular}
\end{center}
\caption{Relative metrics of the online A/B test in JYMBII. }
\vspace{-8pt}
\label{tab.jymbii}
\end{table}

Table \ref{tab.jymbii_segment} shows the A/B test result improvement of segments of members lacking in predictive data after incorporating GNN encoder. In this experiment, we study the model performance on Opportunistic Job Seeker(casually looking) and Urgent Job Seeker(at least 1 apply in past 4 weeks). 
Notable enhancements include a $+3.2\%$ relative increase and a $+2.6\%$ relative rise in Qualified Applications (QA) respectively. 
This improvement demonstrates the GNN model improves the model performance across all the segments. Note that online metrics lift in those segments historically lacking in predictive data is even higher, which is hard to move by any other techniques we have applied so far in LinkedIn's Job Recommender System.

\begin{table}
\begin{center}
\begin{tabular}{@{\hskip 20pt}l@{\hskip 8pt}l}
\toprule
Metric Name &  Impact \\
\midrule
Qualified Applications - Opportunistic     & $+3.2\%$   \\
Qualified Applications - Open to Job       & $+2.8\%$   \\
Qualified Applications - Urgent            & $+2.6\%$   \\
Dismiss To Apply Ratio - Opportunistic     & $-13.8\%$  \\
Dismiss To Apply Ratio - Open to Job       & $-24.2\%$  \\
Dismiss To Apply Ratio - Urgent            & $-25.3\%$  \\
\bottomrule
\end{tabular}
\end{center}
\caption{Relative metrics of the online A/B test for members lacking in predictive data }
\vspace{-8pt}
\label{tab.jymbii_segment}
\end{table}


\subsection{Job Search}
Table \ref{tab.js} presents the key metrics from our experiment incorporating GNN encoder to the Job search ranking product, another key vehicle for Job Marketplace. Compared to the aforementioned JYMBII experiment, this online A/B test demonstrated a greater improvement in job seeker relevancy, evidenced by successful job search sessions and the Apply to Job View ratio. While there was no increase in Qualified Applications, we observed a rise in overall application volume and positive recruiter interactions.

\begin{table}
\begin{center}
\begin{tabular}{@{\hskip 20pt}l@{\hskip 8pt}l}
\toprule
Metric Name &  Impact \\
\midrule
Total Applies          & $+0.5\%$   \\
Apply to view ratio    & $+0.5\%$\\
Successful job search sessions            & $+0.6\%$   \\
Onsite apply positive interactions        & $+0.8\%$   \\
\bottomrule
\end{tabular}
\end{center}
\caption{Relative metrics of the online A/B test in Job Search. }
\vspace{-8pt}
\label{tab.js}
\end{table}

\subsection{Embedding-based Retrieval (EBR)}
Table \ref{tab.ebr} presents the key metrics from our experiment where we incorporated the GNN encoder into Embedding-based Retrieval (EBR) for the candidate generation of Job Search. This integration enhanced the relevancy and precision of job recommendations in both the Organic and Promoted channels via online A/B test.

\begin{table}
\begin{center}
\begin{tabular}{@{\hskip 20pt}l@{\hskip 8pt}l}
\toprule
Metric Name & Impact (Organic) \\
\midrule
Successful Job Search Sessions & $+2.4\%$ \\
Apply Clicks & $+1.5\%$ \\
Apply To Viewport Ratio & $+0.9\%$ \\
CTR & $+1.5\%$ \\

\midrule
Metric Name & Impact (Promoted) \\
\midrule
Successful Job Search Session & $+1.1\%$ \\
Apply Click & $+0.4\%$ \\
Apply To Viewport Ratio & $+0.9\%$ \\
CTR & $+1.8\%$ \\

\bottomrule
\end{tabular}
\end{center}
\caption{Relative metrics of the online A/B test in EBR.}
\vspace{-8pt}
\label{tab.ebr}
\end{table}

\subsection{Ablation Study of Nearline Inference}
Table \ref{tab.nearline} presents a ablation study of adding GNN encoder of job in nearline vs offline in the DNN based L2 ranking model. We observed lift in both JYMBII and Job Search. 

\begin{table}
\begin{center}
\begin{tabular}{@{\hskip 20pt}l@{\hskip 8pt}l}
\toprule
Metric Name &  Impact \\
\midrule
Job Sessions                  & $+0.1\%$   \\
Apply Clicks                  & $+0.6\%$   \\
Successful Job Search Session & $+0.8\%$ \\

\bottomrule
\end{tabular}
\end{center}
\caption{Relative metrics of the online ablation A/B test of Nearline inference. }
\vspace{-8pt}
\label{tab.nearline}
\end{table}

\subsection{Summary of Learning from Experiments}


Our GNN experiments in key job products, like JYMBII, showed that one model enhancement can lead to significant online impacts in A/B tests across various models. Specifically, A/B testing on job ranking models revealed a notable improvement in hiring outcomes—a remarkable feat. Additionally, search relevance saw considerable gains. These advancements benefited a wide user range, from frequent to monthly users, significantly engaging the latter group often challenged by the cold start problem in machine learning.

Moreover, equity improvements via the Job Match feature were not just maintained but boosted through GNN experiments, leading to better hiring results and efficiency for groups with limited predictive data. GNN effectively leveraged its network structure to distribute historical information, significantly aiding these groups.

\section{Conclusion and Future Work}
\label{sec.conclusion}
We have presented a novel framework, LinkSAGE that integrates Graph Neural Networks (GNN) into LinkedIn's large-scale job matching systems, overcoming unique challenges associated with one of the largest professional networks. Our approach utilizes a vast job marketplace graph, the first of its scale in the industry, featuring billions of nodes and edges. This graph is not only expansive but also rich in diversity, encompassing various job-related attributes that form a complex, interconnected network.
A key aspect of LinkSAGE is its training and serving methodology, which effectively combines inductive graph learning with an encoder-decoder GNN model mining a heterogeneous, near real-time evolving graph. This system integrates GNN insights into existing deep neural network models through transfer learning, significantly enhancing performance while minimizing online latency. This eliminates the need for costly real-time GNN infrastructure, marking a first in the realm of large-scale recommender systems.

LinkSAGE has shown remarkable improvements in relevance matching and has been particularly beneficial for diverse user groups, including active job seekers and infrequent visitors.
Importantly, it has significantly improved hiring outcomes and efficiency for groups with limited prediction data.
We have conducted various experiments on LinkedIn Job Marketplace, where multiple practical methods are used for graph definition, online deployment, latency optimization, etc. The deployment of this framework across various LinkedIn products has been validated through multiple online A/B tests, demonstrating its effectiveness in boosting user engagement, relevance matching, and member retention. 
Our work represents a significant stride in the application of GNNs in industrial-scale recommender systems. By addressing key challenges and leveraging the unique properties of GNNs, we have opened new avenues for enhancing job matching services.

Looking ahead, our focus will be on further enhancing the scalability and efficiency of the LinkSAGE. We aim to explore more advanced graph learning algorithms and optimization techniques that can handle even larger datasets and more complex network structures. Another area of interest is the continual improvement of the equity aspect of our model, ensuring that the benefits of GNN are extended to an even broader range of user segments. Additionally, we plan to investigate the potential of integrating other emerging machine learning technologies with our GNN framework to further refine our job recommendation and many other recommender systems in LinkedIn.

\section*{Acknowledgments}
\label{sec.ack}

We would like to thank all our colleagues and collaborators from different organizations and different teams at LinkedIn throughout this project.

\bibliographystyle{ACM-Reference-Format}
\bibliography{main}

\end{document}